# Interior Object Detection and Color Harmonization

Sharmin Pathan
AI Center of Excellence
SAS Institute
sharmin.pathan07@gmail.com

**ABSTRACT**
Confused about renovating your space? Choosing the perfect color for your walls is always a challenging task. One does rounds of color consultation and several patch tests. This paper proposes an AI tool to pitch paint based on attributes of your room and other furniture, and visualize it on your walls. It makes the color selection process easy. It takes in images of a room, detects furniture objects using YOLO object detection. Once these objects have been detected, the tool picks out color of the object. Later this object specific information gets appended to the room attributes (room_type, room_size, preferred_tone, etc) and a deep neural net is trained to make predictions for color/texture/wallpaper for the walls. Finally, these predictions are visualized on the walls from the images provided. The idea is to take the knowledge of a color consultant and pitch colors that suit the walls and provide a good contrast with the furniture and harmonize with different colors in the room. Transfer learning for YOLO object detection from the COCO dataset was used as a starting point and the weights were later fine-tuned by training on additional images.

The model was trained on 1000 records listing the room and furniture attributes, to predict colors. Given the room image, this method finds the best color scheme for the walls. These predictions are then visualized on the walls in the image using image segmentation. The results are visually appealing and automatically enhance the color "look-and-feel".

**Keywords**
YOLO; COCO; Image Segmentation

# 1. INTRODUCTION

Our personal space reflects our personality and the kind of mood we would like to set. The style of this space is highly influenced by our choice of comfort, colors, design, culture. But, along with these choices or preferences, the aesthetics need to blend in. Balance is the key in maintaining aesthetic. Color consultants list out for you the dos and don'ts for achieving that perfect look. Other times people try out patch tests with different paints / textures. Several products offer a tool that allow you to virtually paint your walls by uploading a picture of your room. However, this process is tedious and can be highly confusing with the vast variety of color/texture/wallpaper options available in these tools. At times there are thousands of color choices and one doesn't really go through all of them. The user might land on one color family, experiment a little and might give up later or settle on the best match so far. Even if the user makes a choice, there might be a better choice available that would enhance the look-and-feel and create a pleasant visual perception.

Although there is no set of rules governing the best possible color match given a room or any other space [3], color consultants study color harmony and know how to experiment with these shades and what color family goes along with the other. This basically relies on a consensus among artists that defines when a set is harmonic, and there are some forms, schemes and relations in color space that describe a harmony of colors [5]. Additionally, these professionals rely on experience and intuition to choose the best combination of colors. This paper takes this knowledge of color consultants, their experiences and models it into an AI tool that can help users get an advice on renovating their space, much like a professional color consultant would. This is a novel application of automating the color consulting process along with visualization of the recommendations directly in the user's space. This technique can deal with arbitrarily complex room structures, extracting multiple furniture information, and analyzing a rich variety of colors. It also saves time and effort involved the process. The easy-to-use tool is both suitable for professionals to play around with and provide feedback, also with users who would want to renovate their space.

The tool basically segments the background out of the image, tags it as a wall. It then operates on the foreground, picking up furniture information and different color schemes. Gathering all these inputs and user preferences, it then recommends a color for room walls and visualizes it on the extracted background from the image. Finally, a feedback mechanism gathers inputs from the user and these are used to fine-tune model weights for better predictions in the future.

## 2. BACKGROUND AND RELATED WORK

Several paint companies do provide an interactive tool that lets their customers visualize paint choices by uploading a picture of their room. These tools let the customer experiment with colors or pick the ones already available. But this process can be tedious. The customer doesn't always know what to look for and might get lost in the variety of options in front of them.

Daniel Cohen-Or et al [2] talk about color harmony with the study of physical nature of objects and colors. The paper details how the colors can be picked to maintain an aesthetically pleasing human visual perception. Their algorithm modifies the background colors of an image to harmonize it with the foreground elements. The proposed paper relies on the same ideology. We extract the furniture information using YOLO object detection and use it to determine what colors and color tones would aesthetically enhance the look of the room.

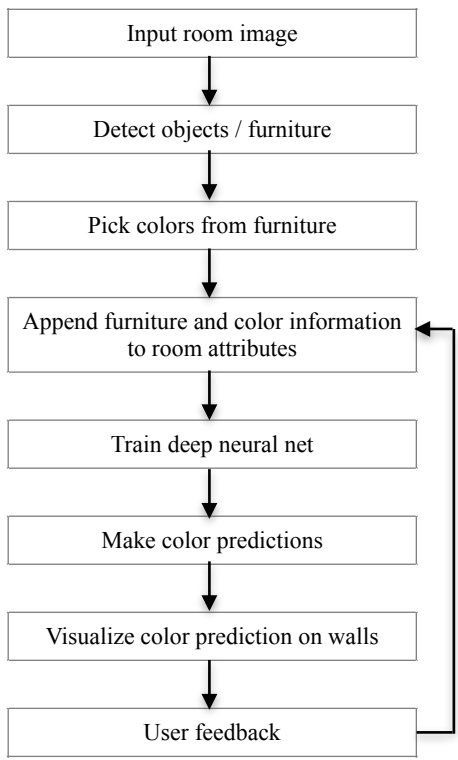

**Figure 1. Model pipeline.**

## 3. MODEL PIPELINE

The end-to-end process has several stages for preparing the data and visualizing color recommendations. An image of the room to be recolored is used as input to be fed into the proposed architecture. It undergoes several stages, the final stage being predicted color visualization on the same image followed by a feedback from user whether they approve of the recommendation. Figure 1 shows the end-to-end process of the pipeline flow. It demonstrates the different stages an input room image goes through. Along the same lines, Figure 2 is an actual visualization of the said pipeline.

### 3.1 YOLO Object Detection

This paper uses YOLO[4] (You Only Look Once) object detection convolution neural net architecture for detecting objects in the scene. The objects here being furniture and the scene being the room image. The object detection stage is built on transfer learning using pre-trained weights from COCO [1] dataset. Classes from the indoor-furniture group are used to detect objects, which include desk, table, bed, couch, chair, furniture-other, cupboard, cabinet, shelf. Additional labeled images were fed to the model to fine tune weights and make better detections.

### 3.2 Identify Dominant Colors

After the objects have been detected, pixel level information is extracted and dominant colors in the identified objects are collected. The pixels are clustered using K-Means clustering and centroids for these color labels are computed. The most popular centroid is picked as the dominant color. RGB values of the dominant color is converted to a decimal value to feed to the deep neural net used in later stages of the pipeline. This information is now organized in a structured format that lists out detected objects and associated colors.

### 3.3 Append Structured Room Attributes

Various attributes associated with the room are collected. This includes the following

- room_type (living_room, bedroom, kitchen)
- room_size (small, medium, big)
- room_style (modern, classic, elegant, traditional)
- room_mood (warm, cool, active, casual, playful)
- room_tone (dark, light, vibrant)

In addition to this, user preferences are also added as attributes and these include

- color_preferences (multiple choices)
- paint_preference (plain_shades, texture, wallpaper)

All these attributes are then appended to the furniture information extracted from user room image.

### 3.4 Train A Deep Neural Net

The room attributes, user preferences and detected furniture are compiled to train a deep neural net model. The target attribute here is color to be recommended for the room. The data is shuffled and split into train, validation, and test sets. 80% of the data goes to training and, 20% for validation. A fully connected artificial neural network consisting of 3 hidden layers is constructed. Each of the hidden layers have

256 neurons and RELU activation. The input layer reads room attributes, user preferences and detected furniture. The model is trained for 200 epochs with a learning rate of 0.01 and uses adam optimization, and dropout of 0.1. The final output layer uses softmax activation and predicts color from the different color families. Table 1 outlines the architecture of the said artificial neural network with the different layers, activation functions and output parameters associated with every layer.

### 3.5 Make Predictions

The color target attributes have been limited to certain color families as every paint company has a certain set of shades to offer. Even if this model can predict a variety of shades to best match the room, what if that shade isn't available at the company? Which is why this paper limits the target attributes to certain color families to demonstrate the effectiveness of this approach. The model was trained on 10 color families. The color families can be easily extended to thousands of colors and can also include textures and wallpaper, making no modification to the training and predictions. Since this paper uses a synthetic dataset, training was performed on limited number of samples. A recommendation of three choices is made by the model by ranking the class probabilities. Top 3 probabilities are selected as the final recommendations.

### 3.6 Image Segmentation

The final stage in the pipeline is to visualize color/ wallpaper/texture predictions on the room's wall. This paper uses simple edge detection and image segmentation to segment out the walls and then replace that segment of the image with model's prediction. That is how the final visualization is done. It sometimes suffers when the colors in the room don't blend with the furniture. Although the segmentation can be misaligned at times, the main focus is to come up with a good choice of recommendation based on the attributes and preferences of the user.

### 3.7 Feedback

Finally, feedback is gathered from user about the prediction. Although the predictions try to maintain aesthetic balance and provide with the best possible recommendation from training and past experiences, it is still the choice of the user. If the user doesn't like the recommendation, they can easily provide a feedback for the same.

This feedback is then incorporated back into the dataset and the model is retrained to fine tune weights and provide with better predictions in the future.

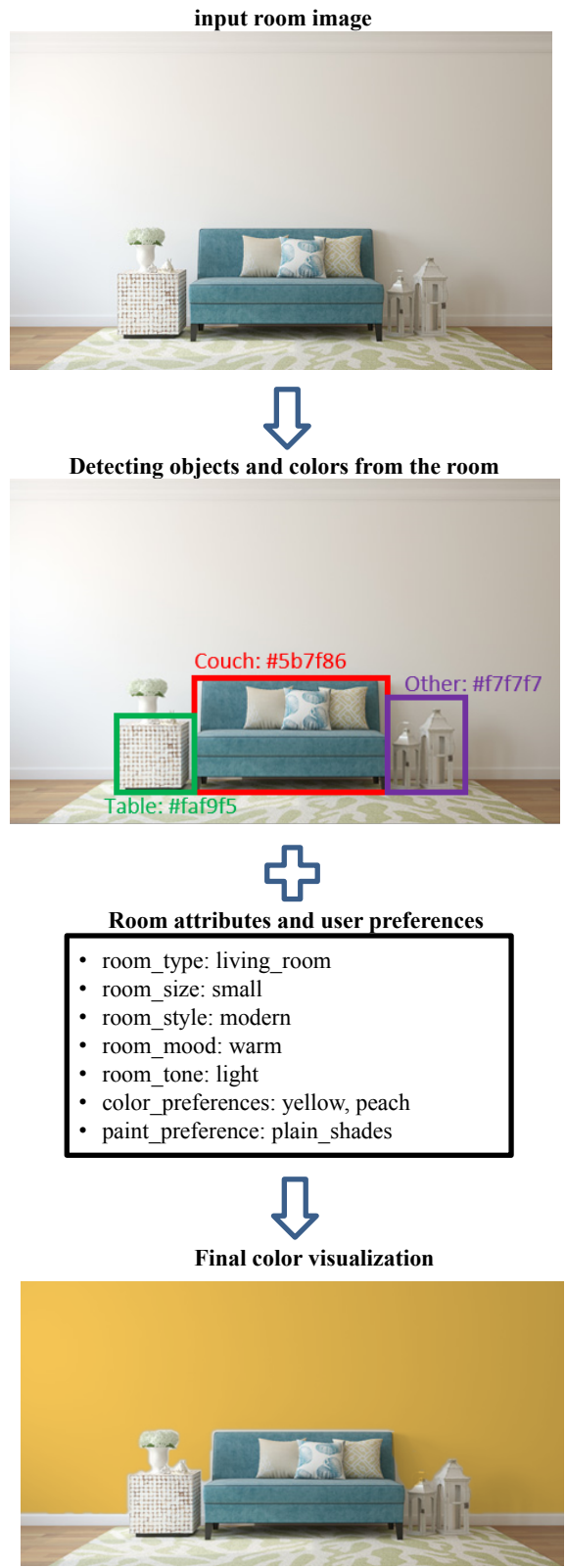

**Figure 2. Model pipeline Visualization.**

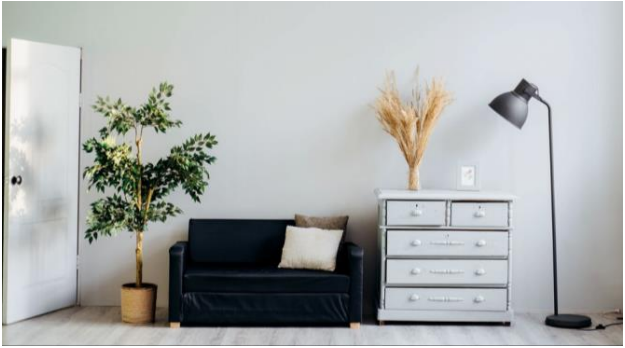
Room 1 input image

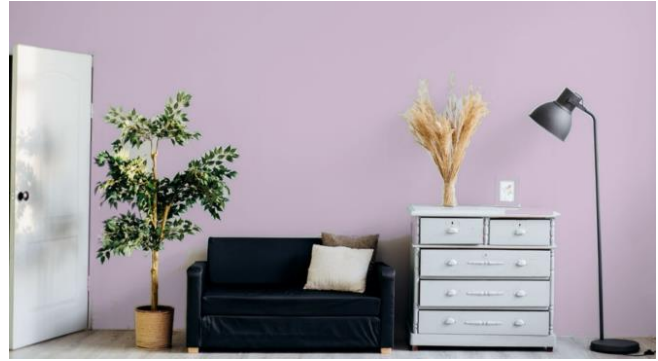
Room 1 color prediction visualization

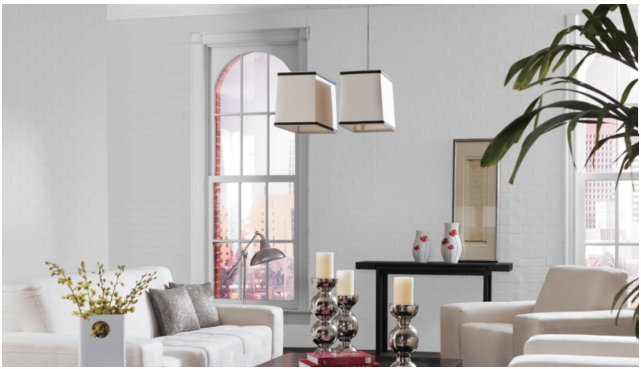
Room 2 input image

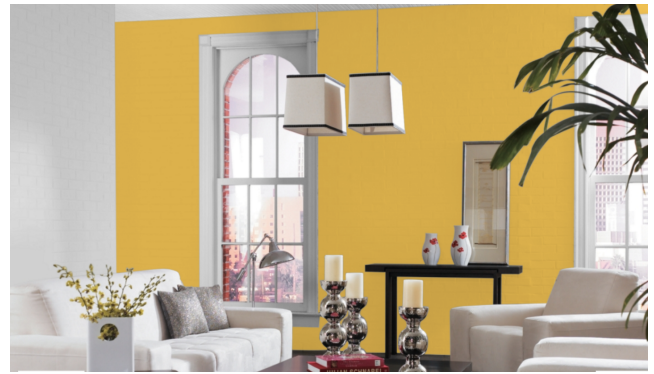
Room 2 color prediction visualization

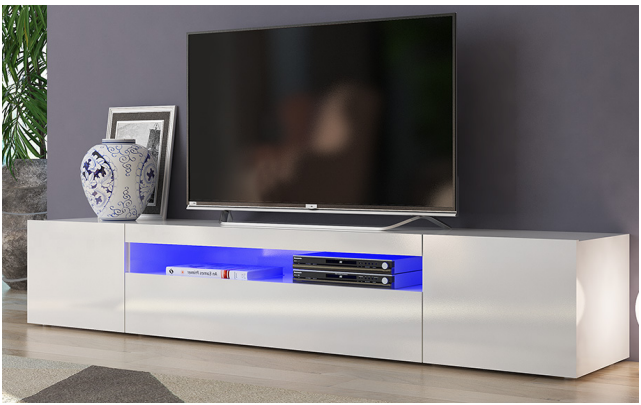
Room 3 input image

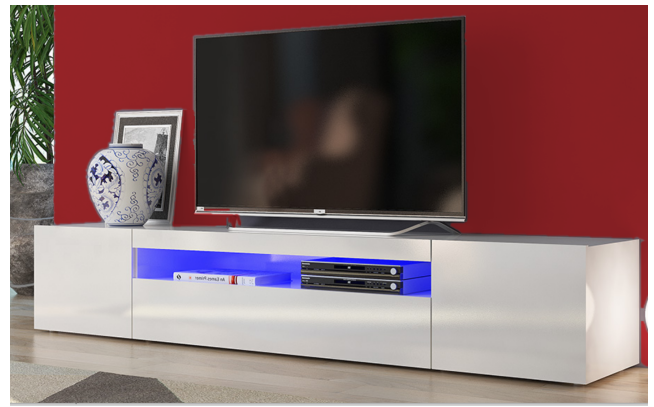
Room 3 color prediction visualization

**Figure 3. Model output visualizations. left: input images, right: model prediction and visualization.**

## 4. RESULTS AND APPLICATIONS

Figure 3 demonstrates results using our approach. It has the input images on the left and final color prediction visualizations on the same image, to the right. With real world data and integrating the tool with a paint industry's offerings, textures and wallpapers can be added as targets to the final predictions. The model ranks top three recommendations for the given space. The visualizations in Figure 3 are for the first recommendation among top 3 recommendations. The model was trained on synthetic data and inputs from a color consultant. Even on synthetic data and limited number of samples, the model performs well and has decent predictions that look acceptable on visualizing them.

## 5. FUTURE WORK

The future work with this architecture is to build a commercially available tool by integrating it with the offerings of a paint industry. The training can be performed on real world data unlike the synthetic data used for the paper. This can be done by gathering insights from color consultants and user inputs/preferences. Also, the feedback mechanism would work better with actual user opinions.

The architecture currently struggles a little with image segmentation stage when the furniture and wall colors just blend in or when there are too many objects cluttered in the room. A better approach can be used for the edge detection based image segmentation to segment out the wall. Moreover, the entire pipeline can serve as a backend to the already available commercial tools.

## 6. CONCLUSION

This paper presents a novel technique of assisting users with paint recommendations using deep learning techniques. These recommendations maintain aesthetic balance by building on the knowledge of color consultants. Currently for the simulated data, records and target colors were limited for training, but these can be easily improved by gathering real world data and customer preferences. Also, a feedback mechanism is incorporated in the pipeline that captures information about whether the user likes the prediction or not. The feedback process can work better when there are multiple users using the tool and providing feedback.

This tool can serve as a great application in paint industry and in households to assist users in selecting plain shades/ wallpapers/textures from the vast variety of available options.